\documentclass[runningheads,a4paper]{llncs}
\usepackage{llncsdoc}
\usepackage{amsmath}
\usepackage{amssymb}
\usepackage{graphicx}
\usepackage[counterclockwise]{rotating}
\usepackage[caption=false]{subfig}
\usepackage{color}
\everymath{\displaystyle\everymath{}}
\usepackage[linesnumbered,ruled]{algorithm2e}
\usepackage{appendix}
\usepackage{hyperref}
\usepackage{scrtime}
\usepackage{marvosym}
\usepackage{multirow}

 \usepackage[space]{grffile} 
 \usepackage{etoolbox} 

\begin{document}
\title{Yet Another ADNI Machine Learning Paper? Paving The Way Towards Fully-reproducible Research on Classification of Alzheimer's Disease}

\author{Jorge Samper-Gonz\'alez$^{1,2}$, Ninon  Burgos$^{1,2}$, Sabrina Fontanella$^{1,2}$, Hugo Bertin$^{3}$, Marie-Odile Habert$^{3}$, Stanley Durrleman$^{1,2}$, Theodoros Evgeniou$^{4}$, Olivier Colliot$^{2,1}$ (\Letter), the Alzheimer's Disease Neuroimaging Initiative}
\institute{
1. INRIA Paris, ARAMIS project-team, 75013 Paris, France
\\
2. Sorbonne Universit\'es, UPMC Univ Paris 06, Inserm, CNRS, Institut du cerveau et la moelle (ICM), AP-HP - H\^opital Piti\'e-Salp\^etri\`ere, 75013 Paris, France
\\
3. Sorbonne Universit\'es, UPMC Univ Paris 06, Inserm, CNRS, LIB, AP-HP, 75013 Paris, France
\\
4. INSEAD, Bd de Constance, 77305 Fontainebleau, France
}
\authorrunning{J. Samper-Gonz\'alez et \emph{al.}}

\titlerunning{Yet Another ADNI Machine Learning Paper?} 

\maketitle

\begin{abstract}

In recent years, the number of papers on Alzheimer's disease classification has increased dramatically, generating interesting methodological ideas on the use machine learning and feature extraction methods. However, practical impact is much more limited and, eventually, one could not tell which of these approaches are the most efficient. While over 90\% of these works make use of ADNI an objective comparison between approaches is impossible due to variations in the subjects included, image pre-processing, performance metrics and cross-validation procedures. In this paper, we propose a framework for reproducible classification experiments using multimodal MRI and PET data from ADNI. The core components are: 1) code to automatically convert the full ADNI database into BIDS format; 2) a modular architecture based on Nipype in order to easily plug-in different classification and feature extraction tools; 3) feature extraction pipelines for MRI and PET data; 4) baseline classification approaches for unimodal and multimodal features. This provides a flexible framework for benchmarking different feature extraction and classification tools in a reproducible manner. Data management tools for obtaining the lists of subjects in AD, MCI converter, MCI non-converters, CN classes are also provided. We demonstrate its use on all (1519) baseline T1 MR images and all (1102) baseline FDG PET images from ADNI 1, GO and 2 with SPM-based feature extraction pipelines and three different classification techniques (linear SVM, anatomically regularized SVM and multiple kernel learning SVM). The highest accuracies achieved were: 91\% for AD vs CN, 83\% for MCIc vs CN, 75\% for MCIc vs MCInc, 94\% for AD-A$\beta$+ vs CN-A$\beta$- and 72\% for MCIc-A$\beta$+ vs MCInc-A$\beta$+. The code is publicly available\footnote{Code at \url{https://gitlab.icm-institute.org/aramislab/AD-ML} - Depends on the Clinica software platform, publicly available at \url{http://www.clinica.run}} .
\end{abstract}

\section{Introduction}

Alzheimer's disease (AD) accounts for over 20 million cases worldwide. Biological and brain imaging markers of AD contain information about complementary processes of the disease progression: in anatomical MRI the atrophy due to gray matter loss, hypometabolism in fluorodeoxyglucose (FDG) PET, the accumulation of amyloid-beta protein in the brain tissue in amyloid PET imaging, as well as concentrations of tau and amyloid-beta proteins in cerebrospinal fluid through lumbar puncture. A major interest is then to analyze those markers to identify AD at an early stage.\par

To this end, a large number of machine learning approaches have been proposed (see \cite{rathore2017review,falahati_multivariate_2014,haller_principles_2011} for a review). Such approaches differ by: i) brain image processing pipelines, 2) feature extraction and selection; 3) machine learning algorithms (classification and/or regression methods). Furthermore, while initial efforts had made use of a single imaging modality (usually anatomical MRI) \cite{kloppel_automatic_2008,fan_spatial_2008,cuingnet_automatic_2011,liu_ensemble_2012,tong_multiple_2014}, a large number of works have proposed to combine multiple modalities (MRI and PET neuroimaging, fluid biomarkers) \cite{zhang_multimodal_2011,young_accurate_2013,gray_random_2013,jie_manifold_2015,teipel_relative_2015,yun_multimodal_2015}. In particular, the combination of anatomical MRI and FDG PET is the most common. \par

Validation and comparison of such approaches require a large number of patients with multimodal data and followed over time. The vast majority (over 90\%) of published works use the publicly available Alzheimer's Disease Neuroimaging Initiative (ADNI). However, objective comparison between their results is impossible because they use: i) different subsets of patients (with unclear specification of selection criteria); ii) different preprocessing pipelines (and thus it is not clear if the superior performance comes for the classification or the preprocessing); iii) different evaluation metrics; iv) different cross-validation procedures. At the end, it is very difficult to conclude which methods perform the best and even if a given modality provides useful additional information. As a result, the practical impact of these works has remained very limited.\par

Comparison papers \cite{cuingnet_automatic_2011,sabuncu_clinical_2014} and challenges \cite{bron_standardized_2015,allen_crowdsourced_2016} have been an important step towards objective evaluation of machine learning methods, by allowing to benchmark different approaches on the same dataset and with the same preprocessing. Nevertheless, such studies provide a "static" assessment of approaches. Evaluation datasets are used in their current state at the time of the study, whereas new patients are continuously included in studies such as ADNI. Similarly, they are limited to the classification and preprocessing methods that were used at the time of the study. It is thus difficult to complement them with new approaches.\par

In this paper, we propose a framework for reproducible evaluation of machine learning algorithms in AD and demonstrate its use on multimodal classification of PET and MRI data of the ADNI database. Specifically, our contributions are three-fold: i) a framework for management of ADNI data and their continuous update with new subjects; ii) a modular set of preprocessing pipelines, feature extraction and classification methods, that provide a baseline for benchmarking of different components; iii) a large-scale evaluation of the added value of the combination of anatomical MRI and FDG PET on 967 patients. By providing a fully automatic conversion into BIDS format, we offer a huge saving of time to users, compared to simply making public the list of used subjects. This is particularly true for complex multimodal datasets like ADNI (with plenty of incomplete data, multiple instances of a given modality and complex metadata). Such a standardized data curation also allows future inclusion of other datasets in a transparent way, perfect for external validation (e.g. the Australian Imaging Biomarker and Lifestyle study, AIBL).\par

Currently, we include relatively standard T1 and (FDG and AV45) PET pipelines and voxel-based SVM classification. The idea is not to impose their use, but to provide a set of baseline tools against which new methods can be easily compared. Researchers working on novel methods can then easily replace a given part of the pipeline (e.g. feature extraction, classification) with their own approach, and evaluate the added value of this specific new component over the baseline approach provided.

\section{Material and methods}

In order to serve methodological developments and to provide a common framework for evaluation and comparison of methods, we developed a unified set of tools for data management, image preprocessing, feature extraction and classification. Data management tools allow to easily update the dataset as new subjects become available in ADNI. Processing and classification tools have been designed in a modular way using Nipype library to allow the development and testing of other methods as replacement for a given step. Then, the impact of each method in the results can be objectively measured. A simple command line interface is provided and the code can also be used as a Python library.

\subsection{Dataset}
ADNI, together with follow ups ADNI GO and ADNI2, is a publicly available database. Overall, it comprises over 300 patients with AD, over 850 patients with mild cognitive impairment (MCI) and over 350 control subjects. An important difficulty with ADNI lies in the organization of the public database. Imaging data, in the state it is downloaded, lacks of a clear structure, and there are multiple image acquisitions for a given visit of a subject. The complementary image information is contained in numerous csv files, making the exploration of the database and subject selection very complicated.

\subsection{A standardized data structure}
To organize the data, we selected the Brain Imaging Data Structure (BIDS)\cite{gorgolewski2016brain}, a community standard to store multiple neuroimaging modalities. Being based on a file hierarchy, rather than a database management system, BIDS can be easily deployed in any environment. For ADNI, among the multiple scans for a visit, we selected a single scan by imaging modality for each subject. In the case of T1 scans, gradwarp and B1-inhomogeneity corrected images were selected when available, otherwise the original image was selected. When repeated MRI were available for a single session, the higher quality scan (as identified in ADNI csv files) was chosen. 1.5T images were preferred for ADNI-1 since they are available for a larger number of patients. For PET scans, the co-registered averaged across time frames images were selected. \par
Very importantly, we provide the code that performs these selections in a fully automated way. The code receives as input a folder containing the download of all ADNI data (in their raw format) and creates a BIDS organized version (T1 MRI, FDG and AV-45 PET). This allows direct reproducibility by other groups without having to redistribute ADNI data, which is not allowed. We also provide tools for subject selection according to desired imaging modalities, time of follow up and diagnose, which make possible to use the same groups with the largest possible number of subjects across studies. Finally, we propose a BIDS-inspired standardized structure for all outputs of the experiments.

\subsection{Preprocessing and feature extraction pipelines}
Two pipelines for processing anatomical T1 MRI and PET images were developed. Pipelines have a modular structure based on Nipype allowing to easily connect and/or replace components. For anatomical T1 MRI, SPM12 was used to perform tissue segmentation (GM, WM, CSF) based on the Unified Segmentation procedure. Next, a DARTEL template is created for all the subjects and a registration to MNI space (DARTEL to MNI) is carried on each of them. The result is that all the images are in a common space, providing a voxel-wise correspondence across subjects. For PET images, we perform a registration of PET data onto the corresponding T1 image in native space. An optional partial volume correction (PVC) step is included using different tissue maps from the T1 in native space.  Then the registration into MNI space using the same transformation as for the corresponding T1 and the generation of a parametric image by normalization to a reference region (the reference region is eroded pons for FDG PET, eroded pons and cerebellum combined for amyloid PET) and masking of non-brain regions are performed.

\subsection{Classification methods}
The obtained images for all the subjects and for each modality lie in a common space so they can be analyzed voxel-wise. The currently implemented classifiers use voxel-as-features (i.e. GM maps for T1 MRI and parametric FDG PET images). Other types of features such as regional features could be easily implemented in the future. We implemented different classifiers for unimodal (T1 MRI or FDG PET) and multimodal classification.

\subsubsection{Unimodal classification.} \textit{Linear SVM.} The first method included to classify single modality images is a linear SVM. For the specified modality, a linear kernel is calculated using the inner product for each pair of images (using all the voxels) for the provided subjects. This kernel is used as input for the generic SVM (makes use of scikit-learn library\footnote{http://scikit-learn.org}). Given its simplicity it is useful as a baseline to compare the performance of methods. \par

\paragraph{Spatially and anatomically regularized SVM.}
The L2 regularization of the standard SVM does not take into account the spatial and anatomical structure of brain images. As a result, the solutions are not easily interpretable (the obtained hyperplanes are highly scattered). To overcome this issue, we implement a classifier that combines a spatial regularization and an atlas based anatomical regularization by using the Fisher metric as proposed by \cite{cuingnet2013spatial}. The spatial proximity takes  into account the distance between voxels while anatomical proximity is defined as if two voxels belong to the same atlas defined region or tissue (in our case GM, WM and CSF probability maps). This approach was initially proposed by \cite{cuingnet2013spatial} in the case of T1 MRI. Here, we extend it to FDG PET images. The implementation of this approach provides a baseline classifier with interpretable maps that can be used for comparison of other methods.


\subsubsection{Multimodal classification.}

For multimodal classification, we included a simple implementation of Multiple Kernel Learning (MKL) for the case of two modalities. A kernel is obtained from the linear combination of two kernels obtained for each imaging modality separately, and then provided as input to the SVM: ~~$K_{\text{T1-FDG}} = \alpha K_{\text{T1}}+(1-\alpha)K_{\text{FDG}}, \quad 0\leq \alpha \leq 1$ \par

Parameter $ \alpha$ that maximizes the balanced accuracy is determined by cross validation. This parameter weights the initial kernels into the combined kernel and therefore represents the contribution of each imaging modality into the classification result.

\subsection{Validation}

\subsubsection{Cross-validation.}
To assess classification performance without bias, it is important to carefully perform cross-validation (CV). In particular, CV must not only concern the training of the classifier, but also the optimization of hyperparameters. While the former is usually dealt with correctly, the latter has not always been appropriately performed in the literature. Here, we implemented two nested CV procedures for hyperparameter optimization and classifier training. By default, a 10-fold CV is implemented in our framework, and an input parameter lets the user select the number of desired folds. 

\subsubsection{Metrics.}
As output of the classification, we report AUC, accuracy, balanced accuracy, sensitivity, specificity and in addition the predicted class for each subject so that the user could calculate any other metric with this information. Also, an image with the SVM weights for each feature is provided. Such weights live in the same space as the MRI and PET images and can thus be used to visualize the brain regions that contributed to the classification.

\section{Results}
We applied the preprocessing pipelines to all (1519) baseline T1 MR images and all (1102) baseline FDG PET images from ADNI 1, GO and 2. Subjects were grouped as AD(239), MCI converter(164), MCI non converter(309) or CN(255), according to diagnosis determined for 36 months of follow up (967 subjects in total). Another grouping was done also taking into account amyloid imaging as AD-A$\beta$+(125), MCIc-A$\beta$+(81), MCIc-A$\beta$-(5), MCInc-A$\beta$+(105), MCInc-A$\beta$-(131) and CN-A$\beta$-(111).

Classification for AD vs CN, MCIc vs CN and MCIc vs MCInc, AD-A$\beta$+ vs CN-A$\beta$- and MCIc-A$\beta$+ vs MCInc-A$\beta$+ classification tasks was performed using voxel-based linear SVM and spatially and anatomically regularized SVM for each modality separately (T1 and FDG PET). Also the multiple-kernel SVM was used on the combination of linear and regularized kernels for T1 and FDG PET obtained in the previous step. Classification results were averaged over 10 runs.
Results can be observed in Table \ref{Results1}. \par
Overall, the use of FDG PET provided better results than T1 in general (balanced accuracy of 91\% for AD vs CN, 83\% for MCIc vs CN, 75\% for MCIc vs MCInc, 94\% for AD-A$\beta$+ vs CN-A$\beta$- and 72\% for MCIc-A$\beta$+ vs MCInc-A$\beta$+), but it significantly outperforms it in the case of conversion prediction (MCIc vs MCInc and MCIc-A$\beta$+ vs MCInc-A$\beta$+ tasks). \par
The combination of different modality kernels confirms this result, providing as best linear combination the case when only FDG PET data was selected, giving no weight to GM kernel.\par
The use of amyloid imaging to refine groups improved the classification results only for FDG PET in the case of AD-A$\beta$+ vs CN-A$\beta$- and not in every task as expected.\par
\begin{figure}[ht!]
\begin{center}
\includegraphics[height=7.5cm]{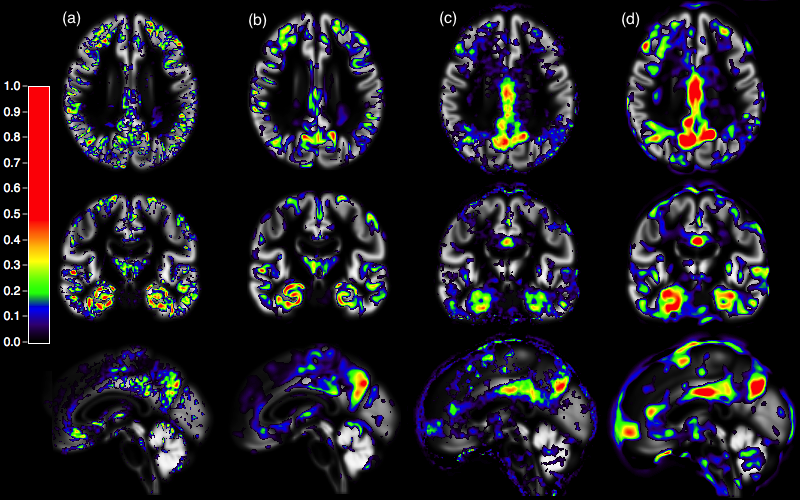}
\caption{Normalized $w^{opt}$ coefficients $\geq$ 0 for AD vs CN: (a) T1 linear SVM, ~~~~~~~~~~~~~~~~~~~~~~~~(b) T1 regularized SVM, (c) FDG PET linear SVM, (d) FDG PET regularized SVM}
\label{reference}
\end{center}
\end{figure}

\section{Conclusions}
We proposed a framework for reproducible machine learning experiments on automatic classification using the ADNI database. It features a standardized data management and a modular architecture for preprocessing, feature extraction and classification. Of note, the dataset can be continuously updated as more data become available. We provide processing pipelines for T1 MRI and PET images as well as classification tools for unimodal and multimodal data.\par

We applied this framework to classify T1 MRI and FDG PET data. We demonstrate that FDG PET imaging provides superior accuracy over T1 MRI when predicting Alzheimer's disease conversion in MCI patients. Furthermore, combination of both modalities did not result in superior accuracy over FDG PET alone. However, more sophisticated combinations may lead to improvements. The use of amyloid imaging to refine groups of subjects didn't improve results in general, but it might be due to the significantly lower subject counts. The classification accuracy for AD vs CN and MCIc vs CN are in line with the state of the art. Nevertheless, there are some papers that report higher  accuracy for prediction of conversion. Our lower accuracy might be due to the fact that we used all available patients, avoiding cherry-picking effects. \par

The code associated to this work will be made available at the time of the conference. We believe that it will be a useful tool for the community to progress towards more reproducible results in this field.

\begin{table}[ht!]
\begin{center}
\begin{tabular}{|c|c|c|c|c|c|c|}\hline
\multirow{2}{*}{\textsc{Image type}} &  \multirow{2}{*}{\textsc{~~Classifier~~}} &  \multirow{2}{*}{\textsc{~~Task~~}}  &  \multirow{2}{*}{\textsc{AUC}} &  {\textsc{~~Bal~~}}  &  \multirow{2}{*}{\textsc{Sens}} &  \multirow{2}{*}{\textsc{Spec}} \\
& & & &\textsc{acc} & & \\ \hline \hline
\multirow{5}{*}{T1} & \multirow{5}{*}{Linear SVM} & AD vs CN & 94.2\%& 88.8\%& 92.8\%& 84.8\% \\  \cline{3-7}
& &  MCIc vs CN  & 85.5\%& 80.5\%& 66.8\%& 89.3\%\\ \cline{3-7}
& &  MCIc vs MCInc  & 73.8\%& 66.5\%& 64.9\%& 69.3\%\\ \cline{3-7}
& &  AD-A$\beta$+ vs CN-A$\beta$-  & 93.6\%& 87.9\%& 90.5\%& 85.5\%\\ \cline{3-7}
& &  MCIc-A$\beta$+ vs MCInc-A$\beta$+  & 73.6\%& 65.8\%& 69.8\%& 60.3\%\\ \hline \hline

\multirow{5}{*}{FDG PET} & \multirow{5}{*}{Linear SVM} & AD vs CN & 96.7\%& 91.1\%& 95\%& 87.2\%\\  \cline{3-7}
& &  MCIc vs CN  & 89.1\%& 83.5\%& 74.7\%& 89.2\%\\ \cline{3-7}
& &  MCIc vs MCInc  & 80.5\%& 75.2\%& 78.5\%& 69.4\%\\ \cline{3-7}
& &  AD-A$\beta$+ vs CN-A$\beta$-  & 98.6\%& 94.4\%& 95.6\%& 93.4\%\\ \cline{3-7}
& &  MCIc-A$\beta$+ vs MCInc-A$\beta$+  & 80.8\%& 72.8\%& 77\%& 67\%\\  \hline

\end{tabular}
\end{center}
  \caption{Classification results (mean of 10 runs)}\label{Results1}
\end{table}

\bibliographystyle{plain} 
\bibliography{biblio} 

\newpage

\section*{Supplementary material}

\begin{table}[ht!]
\begin{center}
\begin{tabular}{|c|c|c|c|c|c|c|}\hline
\multirow{2}{*}{\textsc{Image type}} &  \multirow{2}{*}{\textsc{~~Classifier~~}} &  \multirow{2}{*}{\textsc{~~Task~~}}  &  \multirow{2}{*}{\textsc{AUC}} &  {\textsc{~~Bal~~}}  &  \multirow{2}{*}{\textsc{Sens}} &  \multirow{2}{*}{\textsc{Spec}} \\
& & & &\textsc{acc} & & \\ \hline \hline
\multirow{10}{*}{T1} & \multirow{5}{*}{Linear SVM} & AD vs CN & 94.2\%& 88.8\%& 92.8\%& 84.8\% \\  \cline{3-7}
& &  MCIc vs CN  & 85.5\%& 80.5\%& 66.8\%& 89.3\%\\ \cline{3-7}
& &  MCIc vs MCInc  & 73.8\%& 66.5\%& 64.9\%& 69.3\%\\ \cline{3-7}
& &  AD-A$\beta$+ vs CN-A$\beta$-  & 93.6\%& 87.9\%& 90.5\%& 85.5\%\\ \cline{3-7}
& &  MCIc-A$\beta$+ vs MCInc-A$\beta$+  & 73.6\%& 65.8\%& 69.8\%& 60.3\%\\  \cline{2-7}
& \multirow{5}{*}{Regul SVM} & AD vs CN & 94.9\%& 90\%& 92.5\%& 87.3\% \\  \cline{3-7}
& &  MCIc vs CN  & 86.5\%& 81.2\%& 74.7\%& 85.4\%\\ \cline{3-7}
& &  MCIc vs MCInc  & 74.8\%& 67.3\%& 66.8\%& 68.8\%\\ \cline{3-7}
& &  AD-A$\beta$+ vs CN-A$\beta$-  & 94.9\%& 88.1\%& 92.5\%& 84.2\%\\ \cline{3-7}
& &  MCIc-A$\beta$+ vs MCInc-A$\beta$+  & 73.5\%& 65.4\%& 56.8\%& 76.2\%\\ \hline \hline

\multirow{10}{*}{FDG PET} & \multirow{5}{*}{Linear SVM} & AD vs CN & 96.7\%& 91.1\%& 95\%& 87.2\%\\  \cline{3-7}
& &  MCIc vs CN  & 89.1\%& 83.5\%& 74.7\%& 89.2\%\\ \cline{3-7}
& &  MCIc vs MCInc  & 80.5\%& 75.2\%& 78.5\%& 69.4\%\\ \cline{3-7}
& &  AD-A$\beta$+ vs CN-A$\beta$-  & 98.6\%& 94.4\%& 95.6\%& 93.4\%\\ \cline{3-7}
& &  MCIc-A$\beta$+ vs MCInc-A$\beta$+  & 80.8\%& 72.8\%& 77\%& 67\%\\    \cline{2-7}
& \multirow{5}{*}{Regul SVM} & AD vs CN & 96.7\%& 91.1\%& 94.8\%& 86.9\% \\  \cline{3-7}
& &  MCIc vs CN  & 88.8\%& 83.0\%& 76.5\%& 87.1\%\\ \cline{3-7}
& &  MCIc vs MCInc  & 80.4\%& 73.7\%& 74.5\%& 71.9\%\\ \cline{3-7}
& &  AD-A$\beta$+ vs CN-A$\beta$-  & 98.7\%& 95.4\%& 96.9\%& 93.8\%\\ \cline{3-7}
& &  MCIc-A$\beta$+ vs MCInc-A$\beta$+  & 78.0\%& 71.3\%& 73.5\%& 67.8\%\\ \hline
\end{tabular}
\end{center}
  \caption{Classification results for T1 and FDG PET modalities (mean of 10 runs)}\label{SuppResults1}
\end{table}

\newpage

\begin{table}[ht!]
\begin{center}
\begin{tabular}{|c|c|c|c|c|c|}\hline
\multirow{2}{*}{\textsc{Image type}} &  \multirow{2}{*}{\textsc{~~Classifier~~}} &  \multirow{2}{*}{\textsc{~~Task~~}}  &  {\textsc{~~Bal~~}}  &  \multirow{2}{*}{\textsc{Sens}} &  \multirow{2}{*}{\textsc{Spec}} \\
& & &\textsc{acc} & & \\ \hline \hline\multirow{10}{*}{T1+FDG PET} & \multirow{6}{*}{Linear SVM} & AD vs CN & 90\%& 93\%& 87\%\\  \cline{3-6}
& &  MCIc vs CN  & 81\%& 88\%& 73\%\\ \cline{3-6}
& &  MCIc vs MCInc  & 73\%& 79\%& 68\%\\  \cline{3-6}
& &  AD-A$\beta$+ vs CN-A$\beta$-  & 95\%& 95\%& 94\%\\ \cline{3-6}
& &  MCIc-A$\beta$+ vs MCInc-A$\beta$+  & 73\%& 77\%& 69\%\\ \cline{2-6}
&  \multirow{5}{*}{Regul SVM}  & AD vs CN & 92\%& 95\%& 89\%\\ \cline{3-6}
& &  MCIc vs CN  & 82\%& 87\%& 77\%\\ \cline{3-6}
& &  MCIc vs MCInc & 72\%& 75\%& 69\%\\  \cline{3-6}
& &  AD-A$\beta$+ vs CN-A$\beta$-  & 96\%& 97\%& 95\%\\ \cline{3-6}
& &  MCIc-A$\beta$+ vs MCInc-A$\beta$+  & 72\%& 77\%& 67\%\\ \hline

\end{tabular}
\end{center}
  \caption{Best classification results for MKL combining T1 and FDG PET}\label{SuppResults2}
\end{table}

%
%
%
%
%

\end{document}